\documentclass{article} 
\usepackage{iclr2025_conference}
\iclrfinalcopy

\usepackage{amsmath,amsfonts,bm}

\def\eqref#1{equation~\ref{#1}}

\def\1{\bm{1}}

\DeclareMathAlphabet{\mathsfit}{\encodingdefault}{\sfdefault}{m}{sl}
\SetMathAlphabet{\mathsfit}{bold}{\encodingdefault}{\sfdefault}{bx}{n}

\usepackage{amsmath}
\usepackage{amssymb}
\usepackage{mathtools}
\usepackage{mathrsfs}
\usepackage{hyperref}
\usepackage{amsthm}
\usepackage{cleveref}
\usepackage{url}
\usepackage{comment}
\usepackage{booktabs}
\usepackage{array}
\usepackage{multirow}
\usepackage{enumitem}

\theoremstyle{plain}

\newtheorem{lemma}{Lemma}[section]

\title{Revisiting RaBitQ and TurboQuant: A Symmetric Comparison of Methods, Theory, and Experiments}

\author{Jianyang Gao\\
ETH Zurich\\
\texttt{jianyang.gao@inf.ethz.ch} \\
\And
Yutong Gou, Yuexuan Xu, Jifan Shi\\
Nanyang Technological University\\
\texttt{\{yutong003, yuexuan001, jifan002\}@e.ntu.edu.sg} \\
\And
Yongyi Yang\\
University of Michigan\\
\texttt{yongyi@umich.edu} \\
\And
Shuolin Li\\
Tsinghua University\\
\texttt{sl-li23@mails.tsinghua.edu.cn} \\
\And
Raymond Chi-Wing Wong\\
HKUST\\
\texttt{raywong@cse.ust.hk} \\
\And
Cheng Long\\
Nanyang Technological University\\
\texttt{c.long@ntu.edu.sg} \\
}

\begin{document}

\maketitle

\begin{abstract}
This technical note revisits the relationship between RaBitQ and TurboQuant under a unified comparison framework. We compare the two methods in terms of methodology, theoretical guarantees, and empirical performance, using a reproducible, transparent, and symmetric setup. Our results show that, despite the claimed advantage of TurboQuant, TurboQuant performs worse than RaBitQ in most tested settings of inner-product estimation, nearest-neighbor search and KV cache quantization. We further find that several reported runtime and recall results in the TurboQuant paper could not be reproduced from the released implementation under the stated configuration. Overall, this note clarifies the shared structure and genuine differences between the two lines of work, while documenting reproducibility issues in the experimental results reported by the TurboQuant paper.
\end{abstract}

\section{Introduction}
Vector quantization in high-dimensional Euclidean spaces has become a fundamental problem in modern AI systems, including vector databases and large language model (LLM) serving. 
In these settings, the goal of quantization is to (1) reduce memory usage by compressing high-dimensional vector data; 
(2) reduce computational costs related to the vectors;
and (3) preserve the geometric quantities needed by downstream tasks, especially inner products.

Recently, TurboQuant, first posted on arXiv in April 2025 and later accepted at ICLR 2026~\footnote{\url{https://openreview.net/forum?id=tO3ASKZlok}}, has drawn substantial public attention through claims such as ``at least 6x memory reduction and up to 8x speedup, all with zero accuracy loss''~\citep{turboquant,zandieh2026turboquant}. However, these public-facing claims are framed primarily against an uncompressed baseline, and thus, do not by themselves explain how TurboQuant should be understood relative to prior quantization methods.

One of the most directly relevant prior methods is RaBitQ. The original 1-bit RaBitQ was submitted to SIGMOD 2024 in October 2023, accepted, and posted on arXiv in May 2024~\citep{gao2024rabitq}; its multi-bit extension was posted on arXiv in September 2024 and later accepted at SIGMOD 2025~\citep{gao2025practical}. TurboQuant and RaBitQ are closely related: the two lines overlap in application scenarios, share important method-level structure, and emphasize closely related theoretical guarantees. Yet the TurboQuant paper does not provide an accurate and balanced account of that relationship in three respects. At the method level, RaBitQ is not described or compared with sufficient accuracy. At the theory level, some characterizations of RaBitQ's theoretical guarantees are factually incorrect and unsupported. At the experimental level, the setup used for the RaBitQ baseline is not fully disclosed; subsequent email correspondence reveals that the baseline was run under conditions highly unfavorable to RaBitQ.

This technical note is written to address this gap. Our purpose is to place RaBitQ and TurboQuant within a single comparison framework and to characterize precisely what is shared, what is genuinely different, and what conclusions about theory and experiments are justified under a reproducible, transparent and symmetric standard. The report serves two purposes simultaneously: it provides a citable technical comparison and it offers a clear answer to the broader public question of how RaBitQ and TurboQuant are actually compared to each other. 
Our experimental comparison further shows that the empirical claims in the TurboQuant paper are difficult to reconcile with reproducible evaluations using the released artifacts. 
In inner-product estimation and nearest-neighbor search, in many tested configurations, TurboQuant performs worse than RaBitQ. 
In KV cache quantization, RaBitQ shows clear gains at a bitwidth of 2.5-bit, and the two methods have comparable performance at bitwidth of 3.5-bit.
For quantization time, we find that the reported TurboQuant results cannot be reproduced from the released implementation under the stated hardware configuration, with our measured running times differing substantially from the reported numbers. We also observe inconsistencies between the RaBitQ recall and runtime results reported by the TurboQuant paper and reproduced in this note. These discrepancies indicate that the reported experimental results in the TurboQuant paper are not reproducible from the released artifacts under the stated experimental configuration. In addition, the undisclosed use of asymmetric hardware and parallelism settings for the RaBitQ baseline further weakens the reported comparison as evidence of a consistent empirical advantage over RaBitQ.

We emphasize that this note is not intended to provide a comprehensive survey or benchmark of vector quantization methods in vector databases, LLM serving systems, or broader application domains~\footnote{We recently became aware of related works, DRIVE~\citep{NEURIPS2021_0397758f} and EDEN~\citep{pmlr-v162-vargaftik22a}, which study random rotation followed by quantization for vector reconstruction in the context of federated learning. We include these citations for transparency and to acknowledge related prior work. In this report, we focus on the comparison between TurboQuant and RaBitQ.}. 
It focuses exclusively on the overlap between RaBitQ and TurboQuant and on establishing how they should be compared on common technical ground. Both methods involve several versions of practical implementations; in the technical note, we specify the version in detail for transparency. We also provide code for reproducing all experimental results in this report to facilitate independent evaluation.
Throughout this note, we use RaBitQ to refer to the current RaBitQ line of work, including the original 1-bit RaBitQ~\citep{gao2024rabitq}, its multi-bit extension~\citep{gao2025practical}, its GPU version~\citep{shi2026gpu}, and the optimized implementations in the RaBitQ Library~\citep{gao2025rabitq}.\footnote{\url{https://github.com/VectorDB-NTU/RaBitQ-Library}}

The remainder of this note is organized as follows. We first compare the methodology of both methods under the same comparison framework, followed by a symmetric comparison of their theoretical claims. We next examine the experimental setup and baseline disclosure issues relevant to their reported empirical comparison. We conclude with what we believe is the most accurate characterization of the relationship between the two lines of work.

\section{Comparison of Methodology}

\begin{table}[t]
\centering
\small
\caption{Comparison of RaBitQ and TurboQuant.}
\label{tab:rabitq-vs-turboquant}
\renewcommand{\arraystretch}{1.3}
\begin{tabular}{m{2.0cm}|m{5.5cm}|m{5.5cm}}
                         & \textbf{RaBitQ} & \textbf{TurboQuant} \\ \hline
Preprocessing            & Random rotation/JL Transformation. & Random rotation/JL Transformation. \\ \hline
Codebook                 & Uniform codebook obtained by shifting unsigned integers.
& Non-uniform codebook constructed by $k$-means. \\ \hline
Quantization Algorithm   
& Select a rescaling factor $t$, rescale the rotated and normalized vector, and quantize each coordinate to its nearest codebook entry. A scalar factor is additionally stored for vector reconstruction or inner-product estimation.
& Quantize each coordinate of the rotated and normalized vector to its nearest codebook entry, and store the norm of the input vector. For unbiased inner-product estimation, TurboQuant additionally applies QJL to the residual. \\ \hline
Quantization Code        
& One scalar and $D$ unsigned integers of $B$ bits each.
& One scalar and $D$ unsigned integers of $B$ bits each; for unbiased inner-product estimation, one additional scalar is required. \\
\hline
Inner Product Estimation & Based on native arithmetic of unsigned integers without decoding. & Requires codebook lookup for decoding. \\ \hline
Theoretical Guarantee    
& \textbf{Optimal}: provides a sub-Gaussian tail bound on the error: the required bit-width scales as $\log\log(1/\delta)$.
& \textbf{Suboptimal}: provides a bound on the mean squared error: the required bit-width scales as $\log(1/\delta)$. \\
\end{tabular}
\end{table}

\subsection{Problem Setting}
Both RaBitQ and TurboQuant are generic vector quantization methods for high-dimensional vectors in Euclidean space, aiming to preserve geometric quantities from compressed representations. 
In applications such as approximate nearest neighbor search and LLM systems, a central goal is to preserve inner products between vectors~\footnote{In approximate nearest neighbor search, preserving inner products can also help preserve Euclidean distances, since computing Euclidean distances can be reduced to computing inner products.}. 

A quantization algorithm operates in two stages. In the quantization stage, each input data vector is mapped to a compact representation called a \emph{quantization code}. In the estimation stage, the quantization code is used to estimate the inner product between the data vector and an arbitrary query vector. 

The performance of a quantization algorithm is usually evaluated along five dimensions: (1) quantization time, (2) space consumption, (3) accuracy of inner-product estimation, (4) inner-product estimation time, and (5) theoretical guarantees. Throughout the paper, $\|\cdot\|$ denotes the $\ell_2$ norm unless otherwise specified.

\subsection{Methodology}
In their original papers and artifacts, RaBitQ and TurboQuant are described using different terminology. 
To compare the two methods clearly, we describe the algorithmic procedures of both RaBitQ and TurboQuant under the same framework.
A summary comparison between RaBitQ and TurboQuant is provided in Table~\ref{tab:rabitq-vs-turboquant}.

\subsubsection{Preprocessing of Both RaBitQ and TurboQuant.}

Both RaBitQ and TurboQuant encode the norm and direction of a vector separately. In practice, once the norm is stored, the core quantization procedure of each method focuses on quantizing the normalized vector.

Both RaBitQ and TurboQuant apply a random rotation, which is a form of Johnson--Lindenstrauss Transformation~\citep{johnson1984extensions}, as the first step for all vectors. Both methods use the distributional information of vectors after random rotation to design their quantization algorithms. Specifically, both methods sample and store a random rotation matrix,
and apply the same random rotation to all vectors. In the following sections, without further specification, we assume that all vectors have been rotated by this random matrix.

\subsubsection{Quantization of RaBitQ.}
RaBitQ constructs its codebook from shifted grids of unsigned integers. Let $B$ denote the bit-width per dimension. For an input vector $\mathbf{x}$, RaBitQ first rescales the vector by a factor $t$, then rounds each coordinate of the rescaled vector $t\cdot\mathbf{x}$ to the nearest point in a scalar codebook:
$$
\left\{i - \frac{2^B - 1}{2} \;\middle|\; i = 0, 1, \ldots, 2^B - 1\right\},
$$
and stores the corresponding unsigned integer for each coordinate. 

Across the RaBitQ line and its implementations, three strategies are used to decide the rescaling factor $t$, where the first two strategies decide $t$ on a per-vector basis and the last strategy uses the same $t$ for all vectors:
\begin{itemize}
  \item enumerating all critical rescaling factors that yield distinct quantization codes and selecting the one that maximizes the cosine similarity between the original vector and its quantized counterpart~\citep{gao2025practical};
  \item enumerating candidate rescaling factors from a prescribed set and selecting the one that maximizes the same cosine similarity~\citep{shi2026gpu};
  \item sampling random vectors uniformly from the unit sphere, precomputing the optimal rescaling factor for each, and using the expected value of these optimal factors for fast quantization~\citep{gao2025rabitq}.
\end{itemize}

Let $\mathbf{x}_u \in \{0,1,\ldots,2^B-1\}^D$ denote the vector of $B$-bit unsigned integers produced by the procedure above, and define its shifted-grid representation as
$$
\hat{\mathbf{x}} := \mathbf{x}_u - \frac{2^B - 1}{2}\,\mathbf{1}_D,
$$
where $\mathbf{1}_D$ is the all-ones vector in $\mathbb{R}^D$. The vector $\hat{\mathbf{x}}$ determines the quantized direction; an additional scalar factor is stored to incorporate the norm of the original vector and to support different objectives.

Let $\mathrm{cos}(\mathbf{a},\mathbf{b}) := \left\langle \frac{\mathbf{a}}{\|\mathbf{a}\|},\, \frac{\mathbf{b}}{\|\mathbf{b}\|} \right\rangle$ denote the cosine similarity of two vectors. 
For unbiased inner-product estimation, RaBitQ stores the scalar
$$
\frac{\|\mathbf{x}\|}{\|\hat{\mathbf{x}}\|} \cdot \frac{1}{\mathrm{cos}(\mathbf{x},\hat{\mathbf{x}})}.
$$

We note that while RaBitQ was originally designed for unbiased inner-product estimation, it has also been adapted for vector reconstruction in the RaBitQ library.
Specifically, to instead minimize the reconstruction error, it suffices to replace the scaling factor with
\[
\frac{\|\mathbf{x}\|}{\|\hat{\mathbf{x}}\|} \cdot \cos(\mathbf{x}, \hat{\mathbf{x}})
\]

\subsubsection{Estimation of RaBitQ.}
\label{subsubsec:estimation-rabitq}
Given a query vector, RaBitQ estimates the inner products between the data vectors and the query vector using the quantized representations. Since all data vectors have been rotated, RaBitQ rotates the query vector by the same matrix to preserve inner products; let $\mathbf{y}$ denote this rotated query vector.

RaBitQ estimates the inner product between a data vector $\mathbf{x}$ and the vector $\mathbf{y}$ as follows.
$$
\langle\mathbf{x},\mathbf{y}\rangle
\;\approx\;
\frac{\|\mathbf{x}\|}{\|\hat{\mathbf{x}}\|}\cdot \frac{1}{\mathrm{cos}(\mathbf{x},\hat{\mathbf{x}})}
\cdot
\langle\hat{\mathbf{x}},\mathbf{y}\rangle.
$$
Based on the distribution of vectors after Johnson-Lindenstrauss Transformation, as proved in \citep{gao2024rabitq,gao2025practical}, the above estimator is unbiased and has a rigorous error bound.
The scalar factor $\frac{\|\mathbf{x}\|}{\|\hat{\mathbf{x}}\|}\cdot\frac{1}{\mathrm{cos}(\mathbf{x},\hat{\mathbf{x}})}$ in the estimator is precomputed and stored during the quantization stage. The remaining term $\langle\hat{\mathbf{x}},\mathbf{y}\rangle$ is computed as
$$
\langle\hat{\mathbf{x}},\mathbf{y}\rangle
=
\langle\mathbf{x}_u,\mathbf{y}\rangle
-
\frac{2^B-1}{2}\sum_{i=1}^D \mathbf{y}[i].
$$
where $\mathbf{x}_u$ is the stored $B$-bit code of the data vector. The term $\sum_{i=1}^D \mathbf{y}[i]$ depends only on the rotated query and can be computed once and reused across all data vectors. As a result, RaBitQ computes inner-product estimates directly from the compressed representation, i.e., $\mathbf{x}_u$, without any decoding step.

Furthermore, the structure of RaBitQ's quantization code naturally supports incremental estimation. It can decompose a quantization code into two parts, e.g., the most significant bit and the remaining bits. During estimation, RaBitQ can first produce a coarse estimate of the inner product by accessing only the most significant bit. When higher accuracy is needed, it can access the remaining bits to refine the estimate, which helps significantly speed up the estimation in practice. 

When using RaBitQ for vector reconstruction, based on the precomputed scalar factor $\frac{\|\mathbf{x}\|}{\|\hat{\mathbf{x}}\|}\cdot\mathrm{cos}(\mathbf{x},\hat{\mathbf{x}})$, RaBitQ can reconstruct a vector $\mathbf{x}$ as follows.
\[
\mathbf{x}\approx \frac{\|\mathbf{x}\|}{\|\hat{\mathbf{x}}\|} \cdot \cos(\mathbf{x}, \hat{\mathbf{x}}) \cdot \mathbf{\hat x}
\]

\subsubsection{Quantization of TurboQuant.}
The TurboQuant method includes two variants: one optimized for vector reconstruction and the other for unbiased inner-product estimation.

For vector reconstruction, TurboQuant constructs a scalar codebook according to the Lloyd--Max condition. Specifically, after normalization and random rotation, the coordinates of a rotated vector follow the distribution induced by the uniform spherical measure, as characterized in \citep{uniform_spherical_distribution}. For a target bit-width of $B$, TurboQuant constructs a scalar codebook with $2^B$ centroids by solving the corresponding one-dimensional continuous $k$-means problem under this distribution. Each coordinate is then quantized to the index of its nearest centroid, and the compressed representation stores these centroid indices for all coordinates.
Note that it can compute the reconstructed vector, denoted by $\bar{\mathbf{x}}$, by looking-up the codebook based on the stored indices.

For inner-product estimation, TurboQuant introduces a residual-correction stage. Given a total budget of $B$ bits per coordinate, it first applies $(B-1)$ bits to obtain a reconstruction, denoted by $\bar{\mathbf{x}}$, based on the quantization algorithm for vector-reconstruction, and then computes the residual
$$
\mathbf{r} = \frac{\mathbf{x}}{\|\mathbf{x}\|} - \bar{\mathbf{x}}.
$$
TurboQuant then applies Quantized Johnson--Lindenstrauss (QJL)~\citep{qjl} transform to this residual:
$$
\mathbf{q} = \operatorname{sign}(\mathbf{S}\mathbf{r}),
$$
where $\mathbf{S}$ is a $D\times D$ random Gaussian matrix and $\mathrm{sign}(\cdot )$ is the sign function where $\mathrm{sign}(x)=+1$ if $x\ge 0$ and $\mathrm{sign}(x)=-1$ if $x< 0$. In addition to the first-stage quantization codes and the sign vector $\mathbf{q}$, the quantized representation stores the vector's norm $||\mathbf{x}||$ and the residual norm $\|\mathbf{r}\|$.

\subsubsection{Estimation of TurboQuant.}

Given a query vector, similarly, TurboQuant estimates the inner products between the data vectors and the query vector using the quantized representations. Since all data vectors have been rotated, TurboQuant also rotates the query vector by the same matrix to preserve inner products; let $\mathbf{y}$ denote this rotated query vector.

To estimate the inner products between the data vectors and a query vector, TurboQuant combines the first-stage quantization code (using $(B-1)$ bits per dimension) with a QJL-based estimator of the residual (using 1 bit per dimension)~\citep{qjl} as follows.
$$
\langle\mathbf{x},\mathbf{y}\rangle
\;\approx\;\|{\mathbf{x}}\|\cdot 
\left<
\bar{\mathbf{x}}
+
\sqrt{\frac{\pi}{2}}\cdot\frac{\|\mathbf{r}\|}{D}\mathbf{S}^{\top}\mathbf{q},
\mathbf{y}
\right>
=
\| \mathbf{x}\| \cdot \left<\bar{\mathbf{x}},\mathbf{y}\right>
+
\sqrt{\frac{\pi}{2}}\frac{\| \mathbf{x}\| \cdot \|\mathbf{r}\|}{D}
\left<\mathbf{q},\mathbf{S}\mathbf{y}\right>
$$
where $\bar{\mathbf{x}}$ corresponds to the reconstructed vector based on the quantization codes with $(B-1)$ bits per dimension. This estimator is unbiased as proved in \citep{turboquant}. The first component of the estimator still requires decoding the quantization code through the scalar codebook, while the second component uses the stored sign vector and residual norm to correct the bias.

When using TurboQuant for vector reconstruction, we can reconstruct a vector $\mathbf{x}$ as follows.
\[
\mathbf{x}\approx \|\mathbf{x}\| \cdot \mathbf{\bar x}
\]

\section{Comparison of Theoretical Guarantees}

In this section, we compare in detail the theoretical guarantees of RaBitQ and TurboQuant on inner-product estimation.

We focus on the inner-product-oriented variant of each method, since the reconstruction-oriented variant is optimized for reconstruction error and does not provide unbiased inner-product estimation. Under this scope, both RaBitQ and TurboQuant provide unbiased estimators of the inner product between unit vectors.

We first note that both RaBitQ and TurboQuant are randomized algorithms whose estimation error is a random variable. Rather than providing a deterministic guarantee, both methods can provide a probabilistic guarantee: the additive error of inner product between unit vectors is bounded by $\epsilon$ with probability at least $1-\delta$, where $\epsilon, \delta \in (0,1)$. The key quantity of interest is therefore the trade-off among the error bound $\epsilon$, the failure probability $\delta$, and the bit-width $B$.

In 2017, Alon and Klartag~\citep{alon2017optimal} established the optimal trade-off for approximate inner-product sketches under additive-error guarantees, providing matching upper and lower bounds on the bit-width $B$ required to ensure that the additive error of inner-product estimation between unit vectors is bounded by $\epsilon$ with probability at least $1-\delta$. Specifically, as adapted from the proof of Theorem~4.1 in~\citep{alon2017optimal}, when $\frac{1}{\epsilon^2}\log\frac{1}{\delta} \ge D \ge \log\frac{1}{\delta}$, the optimal bit-width satisfies
$$
B = \Theta\!\left(\log\!\left(\frac{1}{D}\cdot\frac{\log\frac{1}{\delta}}{\epsilon^2}\right)\right).
$$
RaBitQ is proved to match this optimal trade-off; see Theorem~3.2 of~\citep{gao2025practical}. It is worth emphasizing that in the optimal case, the bit-width $B$ grows with $1/\delta$ at the rate of $\log\log(1/\delta)$.

In contrast, TurboQuant provides only a guarantee on the variance of the inner-product estimation error; see Theorem~2 of~\citep{zandieh2026turboquant}. A variance guarantee can be converted into a tail bound via Chebyshev's inequality, which we restate as follows. 

\begin{lemma}[Chebyshev's inequality~\citep{Durrett_textbook}]\label{lem:chebyshev}
Let $X$ be a random variable with mean $0$ and variance $\sigma^2$. Then for any $t > 0$,
$$
\mathbb{P}\!\left\{|X| \geq t\right\} \leq \frac{\sigma^2}{t^2}.
$$
\end{lemma}

However, TurboQuant's theoretical guarantee implies only a suboptimal trade-off between the bit-width $B$ and the failure probability $\delta$.
More precisely, TurboQuant bounds only the variance of the estimator and such a guarantee does not directly yield a sub-Gaussian tail bound.
If one applies Chebyshev's inequality to this variance bound, the resulting dependence requires $B$ to scale as $\log(1/\delta)$.
This is exponentially worse than the $\log\log(1/\delta)$ dependence attained by RaBitQ, which Alon and Klartag~\citep{alon2017optimal} showed to be optimal.

\section{Comparison of Experimental Results}

We follow the TurboQuant paper and evaluate RaBitQ and TurboQuant in the following aspects: namely (1) quantization accuracy, (2) quantization efficiency, (3) nearest neighbor search, and (4) KV cache quantization.
It is important to note that the efficiency of inner-product estimation is also a key evaluation criterion for quantization methods. RaBitQ supports efficient inner-product estimation through a combination of algorithmic and system-level techniques, including bitwise operations~\citep{gao2024rabitq}, FastScan~\citep{fastscanavx2}, and incremental estimation~\citep{gao2025practical}. In contrast, the publicly released TurboQuant codebase only provides a conceptual Python implementation for inner-product estimation, rather than an optimized implementation suitable for efficiency benchmarking. Under these circumstances, a fair empirical comparison of estimation efficiency is not possible. We therefore do not include such efficiency experiments in this report.
Accordingly, we do not include efficiency experiments for nearest neighbor search either because the implementation for efficient nearest neighbor search is also missing from the released code of TurboQuant despite its claims in efficient nearest neighbor search.

For TurboQuant, we use the PyTorch implementation made available on OpenReview~\footnote{\url{https://openreview.net/forum?id=tO3ASKZlok}}.
For RaBitQ, we use by default the C++ implementation open-sourced in the RaBitQ library~\footnote{\url{https://github.com/VectorDB-NTU/RaBitQ-Library}}, with the \texttt{faster\_quant} flag disabled and a random orthogonal matrix for vector rotation (consistent with TurboQuant). We note that faster RaBitQ implementations are available in the same library; we use this configuration for reproducibility, as the TurboQuant paper compares against a Python counterpart of it.

We use a cloud instance with Nvidia A100 GPU (80 GiB VRAM) and 16 VCPUs 
by following the original setup of the papers and a dual-socket server equipped with two Intel Xeon Gold 6418H processors (48 cores / 96 threads in total).
For the reproducibility, we compile all the source codes we use for the experiments in this paper here: \url{https://github.com/VectorDB-NTU/rabitq-turboquant-comparison}.

\subsection{Quantization Accuracy}

Following the TurboQuant paper, we use the DBpedia Entities dataset (1,536-dimensional) and randomly sample 100,000 points as the training set and extract 1,000 distinct entries as the query set. 

In the TurboQuant paper, two versions are provided, namely one denoted by TurboQuant$_{prod}$ for unbiased inner product estimation and the other denoted by TurboQuant$_{mse}$ for vector reconstruction (with minimized MSE).
As discussed in Section~\ref{subsubsec:estimation-rabitq}, RaBitQ can also support inner product estimation and vector reconstruction.
We denote the RaBitQ for unbiased inner product estimation by RaBitQ$_{prod}$ and that for vector reconstruction by RaBitQ$_{mse}$.
We then use the two versions of both methods to quantize the training set and estimate the inner products between the training set and the query set based on the quantization codes of training set. 
We vary the bit widths for quantization and measure the estimation error distributions.
We note that RaBitQ$_{mse}$ was not designed for inner-product estimation; we include it here solely for completeness of the comparison.
In the rest part of this section, unless otherwise specified, RaBitQ refers to RaBitQ$_{prod}$.

\begin{figure}
    \centering
    \includegraphics[width=1.0\linewidth]{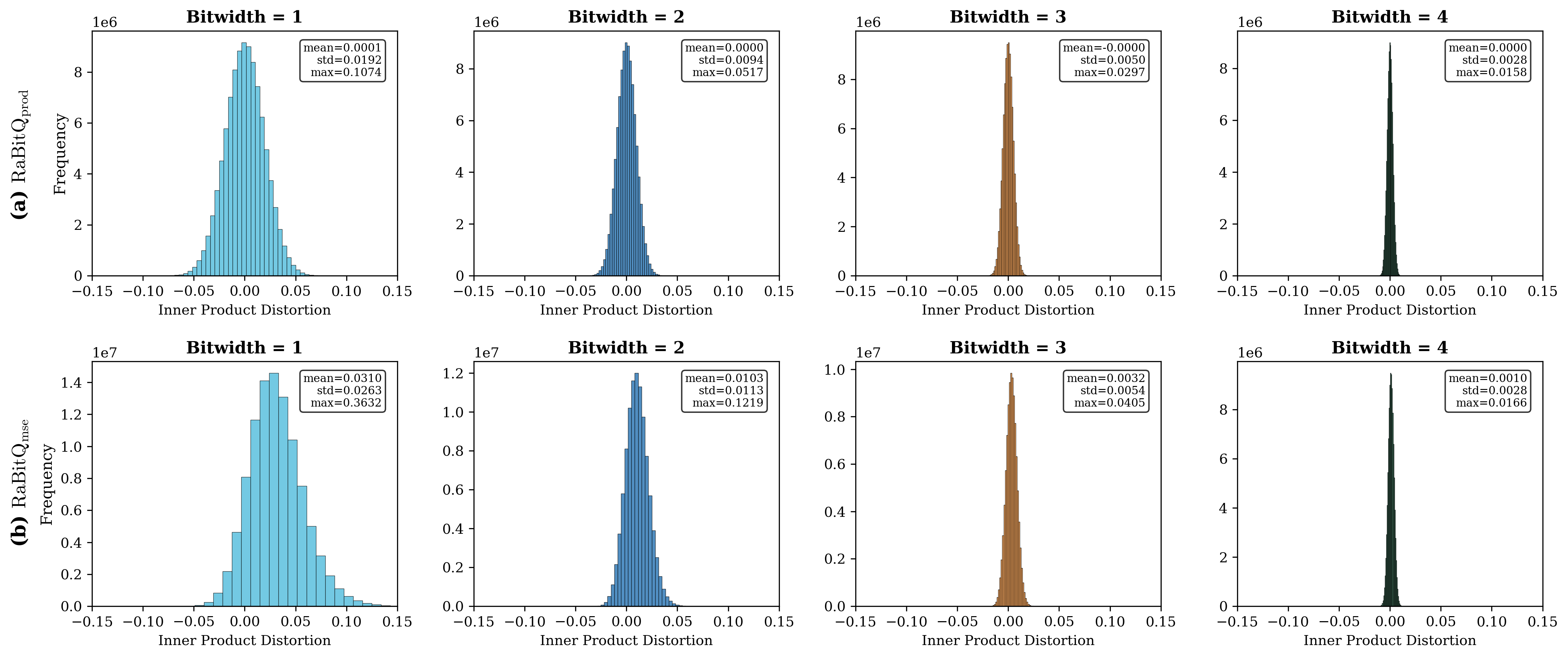}
    \caption{Distribution of Inner Product error for RaBitQ.}
    \label{fig: rabitq ip err fast}
\end{figure}

\begin{figure}
    \centering
    \includegraphics[width=1.0\linewidth]{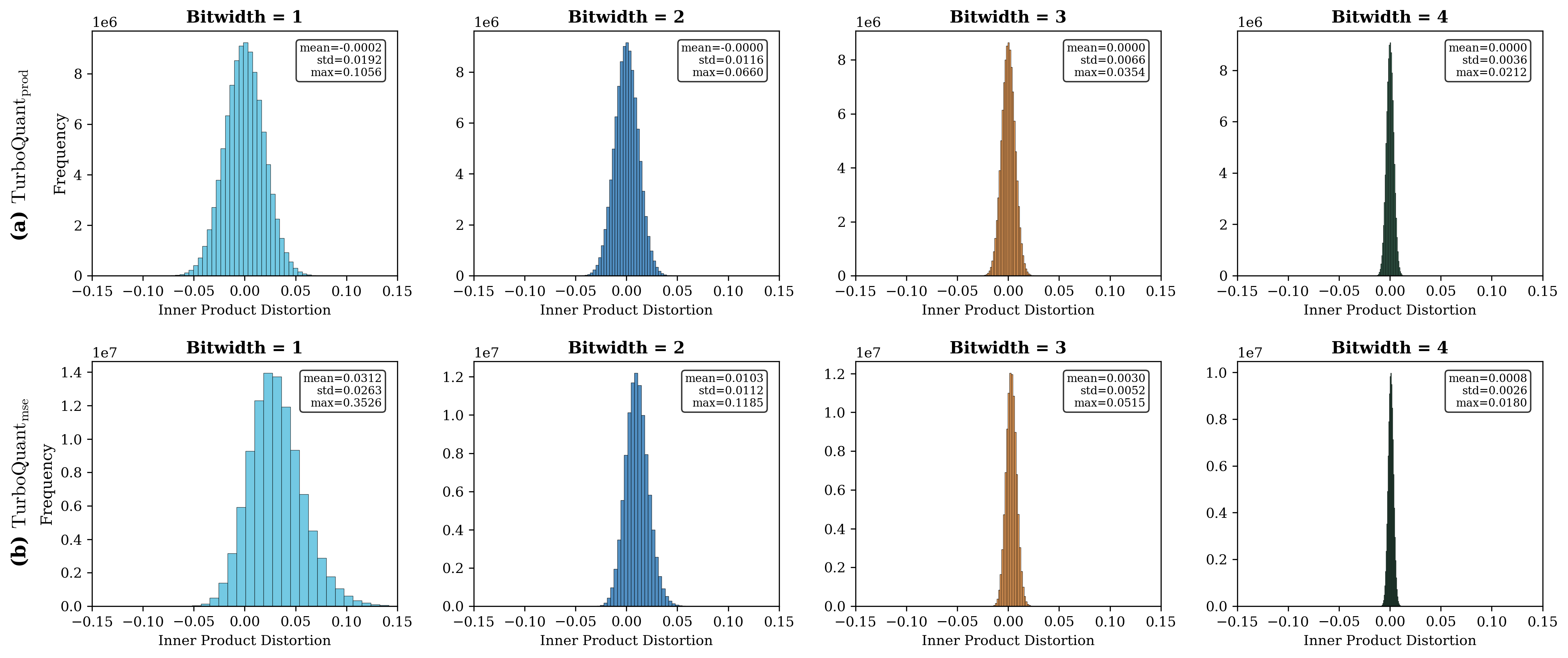}
    \caption{Distribution of Inner Product error for TurboQuant.}
    \label{fig: turbo ip err}
\end{figure}

The results are shown in Figures~\ref{fig: rabitq ip err fast} and~\ref{fig: turbo ip err} for RaBitQ and TurboQuant, respectively. We make the following observations.

\paragraph{Mean error.}
Both RaBitQ$_\mathrm{prod}$ and TurboQuant$_\mathrm{prod}$ maintain a mean error of approximately zero across all bit widths, confirming that both variants are effectively unbiased estimators for inner-product estimation. For the MSE-optimized variants, both methods exhibit a slight positive bias that diminishes as the bit width increases. 

\paragraph{Standard deviation and maximum error.}
For the inner product estimation, RaBitQ$_\mathrm{prod}$ achieves lower standard deviation and maximum error than TurboQuant$_\mathrm{prod}$ at bit widths greater than 1, indicating that RaBitQ produces more tightly concentrated and reliable estimates in the setting where both methods are directly comparable. 

\paragraph{Summary.}
Taken together, these results show that TurboQuant offers no clear and consistent advantage over RaBitQ. In the setting most relevant to inner-product estimation, where the comparison is between the unbiased variants RaBitQ$_\mathrm{prod}$ and TurboQuant$_\mathrm{prod}$, RaBitQ is more stable with smaller std and max errors across most of the tested bit widths.

\subsection{Quantization  Efficiency}

Following the TurboQuant paper,
we use three datasets including GloVe-200 (200-dimensional) and two DBpedia Entities datasets (1536-dimensional and 3072-dimensional).
Specifically, we sample 100,000 vectors from each dataset and quantize the sampled vectors with 4 bits per dimension. 

For RaBitQ, we test four implementations, namely
(1) RaBitQ: which is the default implementation with the \texttt{faster\_quant} flag disabled and a random orthogonal matrix for vector rotation (consistent with TurboQuant),
(2) RaBitQ$_\mathrm{fastOn\text{-}FWHT}$, which is a faster implementation of RaBitQ with the \texttt{faster\_quant} flag enabled and Fast Walsh-Hadamard Transform (FWHT) from the FFHT Library~\citep{ffht_library} and ideas in Kac’s Walk~\citep{kac_walk} for faster vector rotation, (3) RaBitQ (GPU), which is a standalone GPU-based implementation of RaBitQ
and (4) RaBitQ$_\mathrm{fastOn\text{-}FWHT}$ (GPU), which is the GPU version of RaBitQ with the \texttt{faster\_quant} flag enabled and uses FWHT and ideas in Kac’s Walk for faster vector rotation.
For RaBitQ, we use it for reproducibility consideration since the TurboQuant paper implements a Python version of this implementation and uses it for evaluating the performance of RaBitQ.
For RaBitQ (GPU), we use it for a more direct comparison with TurboQuant since the TurboQuant implementation runs on GPU. 
We collect the running time of the methods, with the time for rotating the vectors included.
For GPU-based implementations, the data transfer time from main memory (host memory) to GPU memory (device memory) is excluded.

The results are shown in Table~\ref{tab:quant-time}. We make the following observations.

\paragraph{RaBitQ is faster than TurboQuant on the same hardware.} When compared on the same hardware (i.e., GPU), RaBitQ is substantially faster than TurboQuant. The GPU implementation of RaBitQ outperforms TurboQuant across all three datasets by a large margin: it is approximately $1.2\times$, $1.8\times$, and $1.8\times$ faster at $d = 200$, $d = 1{,}536$, and $d = 3{,}072$, respectively. Moreover, with \texttt{faster\_quant} flag enabled and FWHT, the advances of RaBitQ are more dominant.

\paragraph{RaBitQ on CPU is competitive with TurboQuant on GPU.} Even the CPU implementation of RaBitQ  (RaBitQ$_\mathrm{fastOn\text{-}FWHT}$), which runs on a standard multi-core server without GPU, achieves quantization times within the same order of magnitude as TurboQuant running on an A100 GPU, despite the significant hardware gap. 

\paragraph{Discrepancy with reported RaBitQ results in the TurboQuant paper.} 
The quantization times we observe for RaBitQ differ substantially from those reported in the TurboQuant paper~\citep{turboquant}. This discrepancy is explained by the asymmetric experimental conditions used in the TurboQuant paper. According to our private correspondence with the TurboQuant authors~\footnote{The second author of TurboQuant, Majid Daliri, stated in email correspondence that ``we were using a single-core CPU instance, and multiprocessing was indeed disabled [\ldots] we weren't fully utilizing parallelism, which explains why it was significantly slower''.}, their experiments evaluated RaBitQ on a single-core CPU with multi-threading disabled, while evaluating TurboQuant on an A100 GPU. The TurboQuant paper also implements a Python version of RaBitQ for the evaluation.
These asymmetric setups were not disclosed in the TurboQuant paper.

\paragraph{Discrepancy with reported TurboQuant results in the TurboQuant paper.}
The quantization times we observe for TurboQuant also differ substantially from those reported in~\citep{turboquant}, and the nature of this discrepancy is different. Even when we evaluate TurboQuant using the officially released implementation on the same A100 GPU hardware reported in the paper, we observe quantization times up to approximately two orders of magnitude slower than those reported in~\citep{turboquant}. This suggests that the quantization times reported in the TurboQuant paper are not reproducible from the released implementation under the stated hardware configuration.

\begin{table}[h]
\centering
\renewcommand{\arraystretch}{1.2} 
\begin{tabular}{l rrr}
\toprule
\textbf{Approach} & \textbf{d=200} & \textbf{d=1536} & \textbf{d=3072} \\
\midrule

RaBitQ (CPU) & 0.125 & 1.003 &  4.176 \\
RaBitQ$_\mathrm{fastOn\text{-}FWHT}$ (CPU) & 0.085  & 0.143 & 0.218 \\
RaBitQ (GPU) & 0.009 & 0.065 & 0.152 \\
RaBitQ$_\mathrm{fastOn\text{-}FWHT}$ (GPU) & 0.003 & 0.008 & 0.013\\
TurboQuant (GPU) & 0.011  &  0.114 & 0.276\\

\bottomrule
\end{tabular}
\caption{Quantization time (in seconds) for different approaches across various dimensions using 4-bit quantization.}
\label{tab:quant-time}
\end{table}

\subsection{Nearest Neighbor Search}

Following the TurboQuant paper, we use three datasets, namely GloVe-$200$, OpenAI3-$1536$, and OpenAI3-$3072$. 
For each dataset, we construct a base set and a query set.
For OpenAI3-$1536$ and OpenAI3-$3072$, the base set has $100{,}000$ vectors and the query set contains $1{,}000$ vectors. 
For GloVe-$200$, we sample a subset of $100{,}000$ vectors from the original corpus as the base set and use the provided query set of $10{,}000$ vectors. 
Following the TurboQuant paper, for all three datasets, we use the inner product of normalized vectors as the metric for nearest neighbor search.

We compare RaBitQ, namely RaBitQ$_{prod}$, with the two TurboQuant variants, namely TurboQuant$_{prod}$ and TurboQuant$_{mse}$.
Note that we exclude RaBitQ$_{mse}$ from this comparison as it is not designed for inner-product estimation, which is the objective underlying nearest neighbor search.
On the other hand, we include both variants of TurboQuant in the comparison for transparency as the TurboQuant paper did not specify the version they used in the experiment.
For each method, we first quantize the vectors in the base set and then find for each query vector in the query set, the $k$ vectors, whose quantized vectors have the largest estimated inner products with the query vector.
We vary the bit-width $B$ in $\{2,4\}$.
We report \emph{Recall@1@}$k$ for $k \in \{1,2,4,8,16,32,64\}$. Let $g(\mathbf{q})$ denote the exact top-1 nearest neighbor of query $\mathbf{q}$ (i.e., the one with the largest inner product), and 
let $A_k(\mathbf{q})$ denote the approximate top-$k$ result set returned by a method. Then
\[
\mathrm{Recall@1@}k
=
\frac{1}{|Q|}
\sum_{\mathbf{q} \in Q}
\mathbf{1}\!\left[g(\mathbf{q}) \in A_k(\mathbf{q})\right].
\]

Note that in the open-sourced code of TurboQuant, the evaluation script for the two OpenAI datasets is available, but that for the GloVe-$200$ data is not. 
Therefore, for the OpenAI datasets, we use the provided evaluation script directly; 
and for GloVe-$200$, we use a thin wrapper that calls the same TurboQuant core routines for random rotation generation and quantization; thus, the underlying TurboQuant quantizer itself is unchanged.

In addition, we note that both RaBitQ and TurboQuant involve randomness through their sampled rotation matrices. As a result, recall curves from a single run may exhibit mild run-to-run variation. To obtain a more stable comparison, we repeat each configuration, defined by method, bit-width, and dataset, 10 times using the full query set. We plot the mean recall over these runs as the main curve, and use the shaded band to represent one standard deviation around the mean.

\begin{figure}
    \centering
    \includegraphics[width=1\linewidth]{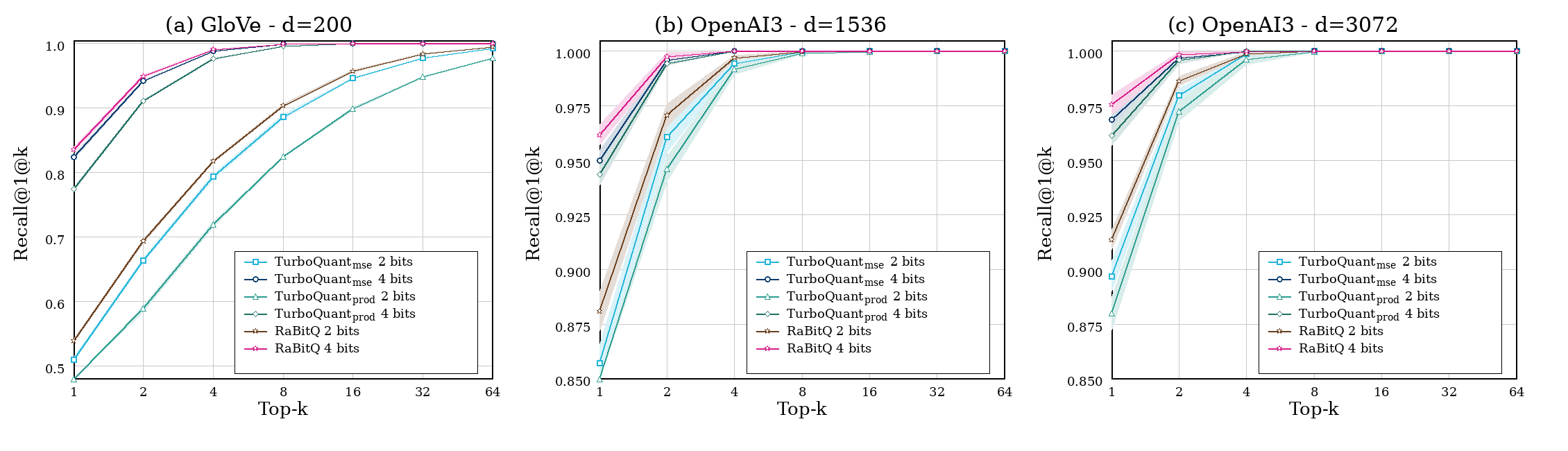}
    \caption{Recall comparison on different datasets}
    \label{fig:recall comparison}
\end{figure}

The results are shown in Figure~\ref{fig:recall comparison}. We make the following observations.

\paragraph{Overall comparison.}
Across all three datasets and both bit widths, RaBitQ consistently achieves higher recall than both TurboQuant variants. The advantage is most pronounced at small $k$ and at the lower bit width of 2 bits, where the methods are most differentiated. 
As $k$ increases, all methods converge toward perfect recall and the differences diminish accordingly.

\paragraph{TurboQuant$_\mathrm{mse}$ outperforms TurboQuant$_\mathrm{prod}$ on recall.}
We observe that TurboQuant$_\mathrm{mse}$ consistently achieves higher recall than TurboQuant$_\mathrm{prod}$ across all settings. This is a notable finding because TurboQuant$_\mathrm{prod}$ is the variant specifically designed for inner-product estimation, which is the objective directly relevant to nearest neighbor search. The fact that the reconstruction-oriented variant yields better recall performance raises questions about which variant should be used in practice for this task, and about the theoretical guarantees that support TurboQuant$_\mathrm{prod}$ in this setting. We note that TurboQuant$_\mathrm{mse}$ does not guarantee unbiased inner-product estimation.
The TurboQuant paper does not clearly specify which variant is used in its reported recall results.

\paragraph{Discrepancy with results reported in the TurboQuant paper.}
We note that the recall values we obtain for RaBitQ differ from those reported in the TurboQuant paper~\citep{turboquant}. Specifically, the RaBitQ results reported therein fall below the one-standard-deviation band we measure across 10 repeated runs, each using the full query set, with different random seeds. The TurboQuant paper does not describe how run-to-run variation due to random rotation is handled in their reported RaBitQ results, making it difficult to assess the source of this discrepancy. Our results, by contrast, are averaged over 10 independent runs with standard deviations reported, and are fully reproducible from the code provided in our repository. These reproduced results do not support the TurboQuant paper’s conclusion that TurboQuant consistently outperforms RaBitQ in nearest neighbor search.

\subsection{KV Cache Quantization}

We compare RaBitQ with TurboQuant for KV cache quantization in long-context generation. We note that the TurboQuant paper does not specify which of its two variants was used for KV cache quantization. Moreover, the open-source community has observed that the QJL-based variant (TurboQuant$_\mathrm{prod}$) can hurt attention quality by amplifying variance through the softmax operation.\footnote{\url{https://docs.vllm.ai/en/v0.20.0/api/vllm/model_executor/layers/quantization/turboquant/}} We therefore use TurboQuant$_\mathrm{mse}$ for this comparison.
 
The released code of TurboQuant contains the core quantization routines (including \texttt{TurboSketch}, centroid tables, and outlier separation), an attention layer that hardcodes TurboQuant as the only backend, and a LongBench evaluation script. However, the code cannot be executed as released: it depends on an unpublished CUDA kernel package and contains multiple bugs in the quantization pipeline.
For example, the value-cache quantizer is never constructed, and the decode-phase quantization logic is unreachable due to an early return. We fix these bugs and provide a unified KV cache framework in which RaBitQ and TurboQuant share identical cache logic, buffer management, and outlier handling, while retaining the original MSE-based centroids. Full details are available in our released code.
In our evaluation, the quantization method is the only varying factor in the KV cache framework, while all other configurations are kept identical.

For a direct comparison, both methods use the same outlier-aware key-cache
bit allocation. For each attention head with $d_h = 128$, the 32 key
channels with the largest L2 norm are quantized at a higher bitwidth, while
the remaining 96 channels use a lower bitwidth. Each quantized key vector
stores two additional float16 scaling values, namely one for the sub-vector consisting of outlier channels and the other for the sub-vector consisting of remaining channels. We evaluate two key-cache
configurations: \textbf{2.5-bit}, using 3-bit outlier channels and 2-bit non-outlier channels, and \textbf{3.5-bit}, using 4-bit outlier channels and
3-bit non-outlier channels, corresponding to effective bitwidths of
$(32 \times 3 + 96 \times 2 + 2 \times 16) / 128 = 2.5$ and
$(32 \times 4 + 96 \times 3 + 2 \times 16) / 128 = 3.5$, respectively.
Values are quantized uniformly at 2 bits, so the 2.5-bit and 3.5-bit labels
refer to the key-cache configuration.

For LongBench-E, we use the official task metrics with the same output
post-processing convention as the released TurboQuant evaluation code.

\paragraph{Needle-In-A-Haystack.}
We evaluate retrieval behavior on Llama-3.1-8B-Instruct using
Needle-In-A-Haystack across 15 context lengths (4k--104k tokens) and 10
needle depths, yielding 150 test points per method.
The released TurboQuant code does not include an NIAH evaluation script.
We build our evaluation on the official
LLMTest\_NeedleInAHaystack~\citep{kamradt2023needle} framework.
However, its default GPT-3.5-turbo judge produces inconsistent scores
for the same model output across repeated evaluations, making results
difficult to reproduce.
We therefore replace it with the keyword-coverage scorer used by
Token-Sparse-Attention~\citep{jo2026token}, which measures
the fraction of expected-answer words that appear in the model
output, yielding a deterministic metric in $[0,1]$.
One additional issue is that the NIAH framework constructs haystacks by
concatenating Paul Graham essays loaded via \texttt{glob.glob}, whose
iteration order is filesystem-dependent and therefore
non-deterministic. 
We record the glob ordering observed on
our machine and provide it in the released code for reproducibility.

\begin{figure}
    \centering
    \includegraphics[width=1\linewidth]{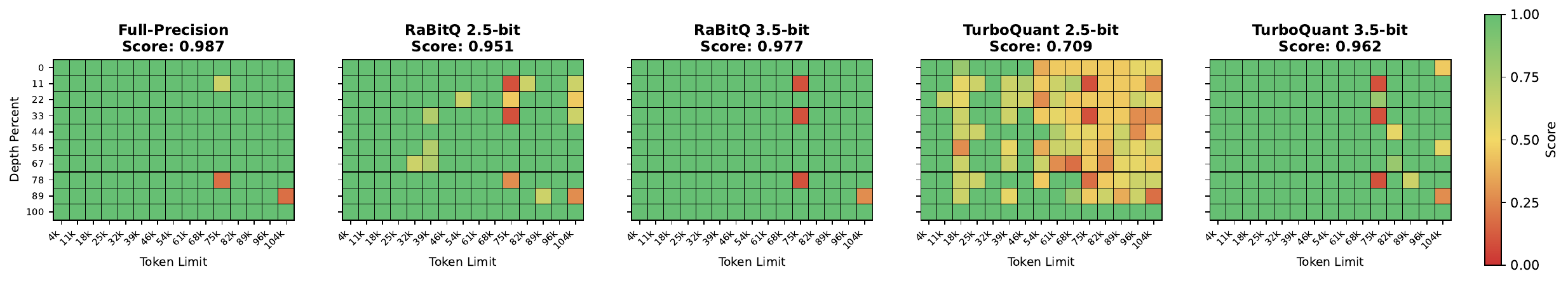}
    \caption{Evaluation of Llama-3.1-8B-Instruct on the “Needle-In-A-Haystack” test}
    \label{fig:needle}
\end{figure}

Results are shown in Figure~\ref{fig:needle}. The full-precision
baseline scores $0.987$. RaBitQ remains close to this level at both
2.5-bit and 3.5-bit, scoring $0.951$ and $0.977$, respectively.
$\text{TurboQuant}_{\text{mse}}$ also performs well at 3.5-bit
($0.962$), but drops to $0.709$ at 2.5-bit: 86 out of 150 test points
score below $0.8$. 
The failures are widespread across nearly all needle
depths (only depth${}=100\%$ is fully correct) and concentrate at
longer contexts, where the mean score falls from $0.898$ ($\le$32k) to
$0.615$ ($>$32k). This suggests that the
MSE-based centroid placement, while adequate at higher bitwidths,
introduces sufficient approximation error at 2.5-bit to distort
attention scores over long sequences, causing the model to fail to
attend to the relevant passage.

\paragraph{LongBench-E.}

We evaluate on all 13 datasets of LongBench-E~\citep{bai2024longbench} using Llama-3.1-8B-Instruct and Ministral-8B-Instruct-2410, grouped into 6 categories. We note that the TurboQuant paper reports results for ``Ministral-7B-Instruct'', which does not correspond to any model available on public model hubs. It might mean Mistral-7B-Instruct, but the model exists in three versions and the paper does not specify which was used. Therefore, we adopt the unambiguous Ministral-8B-Instruct-2410 instead. Category scores are computed as the mean of the dataset-level scores within each category. Following the TurboQuant reporting convention, the overall average is computed over all 13 dataset-level scores rather than over the 6 category scores. 
We follow the TurboQuant paper and explore the bitwidths of 2.5 bits and 3.5 bits on Llama and 2.5 bits on Ministral.

Table~\ref{tab:longbench} shows the same trend at 2.5-bit: RaBitQ achieves
higher average scores than $\text{TurboQuant}_{\text{mse}}$ on both models,
with 48.64 vs.\ 47.78 on Llama-3.1-8B and 52.60 vs.\ 51.80 on
Ministral-8B. The largest category-level gains appear on Code (+2.20 on
Llama and +1.16 on Ministral), where generation depends strongly on
long-range contextual consistency. At 3.5-bit on Llama-3.1-8B, the two
methods are comparable, both close to the full-cache
baseline of 50.39. Overall, RaBitQ shows clearer gains at 2.5-bit, while
the two methods become comparable as the bitwidth increases.

\begin{table}[t]
\centering
\caption{LongBench-E results for Llama-3.1-8B-Instruct and Ministral-8B-Instruct.}
\label{tab:longbench} 
\resizebox{\textwidth}{!}{%
\begin{tabular}{llccccccc}
\toprule
\textbf{Model} & \textbf{Method} & \textbf{SingleQA} & \textbf{MultiQA} & \textbf{Summ} & \textbf{Few shot} & \textbf{Synthetic} & \textbf{Code} & \textbf{Avg} \\
\midrule
\multirow{5}{*}{Llama-3.1-8B}
& Full Cache (16-bit)       & 45.39 & 45.76 & 26.38 & 68.60 & 59.12 & 48.00 & 50.39 \\
& RaBitQ 2.5-bit            & \textbf{43.74} & \textbf{45.49} & \textbf{22.56} & \textbf{67.38} & \textbf{58.96} & \textbf{44.34} & \textbf{48.64} \\

& $\text{TurboQuant}_{\text{mse}}$ 2.5-bit        & 42.20 & 44.36 & 21.89 & 67.37 & 58.94 & 42.14 & 47.78 \\
 & RaBitQ 3.5-bit            & \textbf{44.85} & 45.56 & 24.70 & 67.90 & \textbf{59.53} & \textbf{45.58} & 49.55 \\
& $\text{TurboQuant}_{\text{mse}}$ 3.5-bit        & 44.11 & \textbf{45.75} & \textbf{25.17} & \textbf{68.11} & 59.49 & 45.53 & \textbf{49.57} \\
\midrule
\multirow{3}{*}{Ministral-8B}
& Full Cache (16-bit)       & 51.29 & 57.28 & 25.92 & 69.37 & 58.16 & 56.10 & 54.28 \\
& RaBitQ 2.5-bit            & \textbf{49.39} & \textbf{56.39} & \textbf{22.90} & \textbf{68.71} & \textbf{58.00} & \textbf{52.16} & \textbf{52.60} \\
& $\text{TurboQuant}_{\text{mse}}$ 2.5-bit        & 47.86 & 55.58 & 21.30 & 68.64 & \textbf{58.00} & 51.00 & 51.80 \\
\bottomrule
\end{tabular}%
}
\end{table}
\section{Conclusion}

This note has examined the relationship between RaBitQ and TurboQuant across three
dimensions: methodology, theoretical guarantees, and empirical performance.

At the method level, both RaBitQ and TurboQuant apply a random rotation as their
first step and exploit the resulting distributional properties to design their respective
quantization schemes as well as analyzing the unbiasedness and error bounds for inner product estimation.

At the theoretical level, RaBitQ provably achieves the asymptotically optimal
space-distortion trade-off established by Alon and Klartag~\citep{alon2017optimal},
with a bit-width that grows with the failure probability $\delta$ at the rate of
$\log\log(1/\delta)$. TurboQuant, by contrast, provides only a variance guarantee
on its estimator. Converting this to a tail bound via Chebyshev's inequality yields
a dependence that grows as $\log(1/\delta)$, which is exponentially worse than the
optimal rate.

At the experimental level, our reproducible evaluation shows that TurboQuant offers
no clear and consistent advantage over RaBitQ in directly comparable settings. In
quantization accuracy, RaBitQ$_\mathrm{prod}$ matches or outperforms
TurboQuant$_\mathrm{prod}$ across all tested bit widths. In quantization efficiency,
RaBitQ is substantially faster than TurboQuant on the same hardware, and its CPU
implementation is competitive with TurboQuant on an A100 GPU. In nearest neighbor
search, RaBitQ consistently achieves higher recall than both TurboQuant variants
across all datasets and bit widths. 
In KV cache quantization, RaBitQ shows clear gains at 2.5-bit, and the two methods have comparable performance at 3.5-bit.
Furthermore, we find that the runtime and recall
results reported in the TurboQuant paper could not be reproduced from the released
implementation under the stated hardware configuration.

We hope this note serves as a useful and citable reference for researchers working
on vector quantization, and that the symmetric comparison framework presented here
contributes to a more accurate understanding of the relationship between the two
methods.



\bibliography{iclr2025_conference}
\bibliographystyle{iclr2025_conference}


\end{document}